\documentclass[twocolumn]{asme2e}
\usepackage[dvips]{graphicx}

\newcommand{\be}{\begin{equation}}
\newcommand{\ee}{\end{equation}}
\newcommand{\beq}{\begin{equation}}
\newcommand{\eeq}{\end{equation}}
\newcommand{\bed}{\begin{displaymath}}
\newcommand{\eed}{\end{displaymath}}
\newcommand{\beqa}{\begin{eqnarray}}
\newcommand{\eeqa}{\end{eqnarray}}
\newcommand{\beqann}{\begin{eqnarray*}}
\newcommand{\eeqann}{\end{eqnarray*}}
\newcommand{\bseq}{\begin{subequation}}
\newcommand{\eseq}{\end{subequation}}

\newcommand{\ba}{\begin{array}}
\newcommand{\ea}{\end{array}}

\newcommand{\negr}[1]{{\bf {#1}}}

\def\confshortname{DETC'99}
\def\conffullname{1999 ASME Design Engineering Technical Conferences}
\def\confdate{12-15}
\def\confmonth{September}
\def\confyear{1999}
\def\confcity{Las Vegas}
\def\confstate{Nevada}
\def\confcountry{USA}

\newtheorem{Def}{Definition}

\def\papernum{DETC99/DAC-8645}

\title{Regions of Feasible Point-to-Point Trajectories in the
Cartesian Workspace of Fully-Parallel Manipulators}

\author{Damien Chablat
    \affiliation{
        INRIA Rocquencourt\\
        Domaine de Voluceau, B.P. 105\\
        78 153 Le Chesnay France \\
        Email: Chablat@cim.mcgill.ca
    }
}

\author{Philippe Wenger
    \affiliation{Institut de Recherche en Cybern\'etique de Nantes \\
        1, rue de la No\"e, \\
        44321 Nantes, France \\
        Email: Philippe.Wenger@lan.ec-nantes.fr
    }
}
\begin{document}

\maketitle

\begin{abstract}
The goal of this paper is to define the n-connected regions in the Cartesian workspace of fully-parallel manipulators, i.e. the maximal regions where it is possible to execute point-to-point motions. The manipulators considered in this study may have multiple direct and inverse kinematic solutions. The N-connected regions are characterized by projection, onto the Cartesian workspace, of the connected components of the reachable configuration space defined in the Cartesian product of the Cartesian space by the joint space. Generalized octree models are used for the construction of all spaces. This study is illustrated with a simple planar fully-parallel manipulator.
\end{abstract}
\section*{INTRODUCTION}
The Cartesian workspace of fully-parallel manipulators is generally
defined as the set of all reachable configurations of the moving
platform. However, this definition is misleading since the
manipulator may not be able to move its platform between two
prescribed configurations in the Cartesian workspace. This feature
is well known in serial manipulators when the environment includes
obstacles \cite{Wenger:91}. For fully-parallel manipulators,
point-to-point motions may be infeasible even in obstacle-free
environments. For manipulators with one unique solution to their
inverse kinematics (like Gough-platforms), one configuration of the
moving platform is associated with one unique joint configuration
and the connected-components of the singularity-free regions of the
Cartesian workspace are the maximal regions of point-to-point
motions \cite{Chablat:98}. Unfortunately, this result does not hold
for manipulators which have multiple solutions to both their direct
and inverse kinematics. For such manipulators which are the subject
of this study, the singularity locus in the Cartesian workspace
depends on the choice of the inverse kinematic solution
\cite{Chablat:97} and the actual reachable space must be firstly
defined in the Cartesian product of the Cartesian space by the
joint space. The goal of this paper is to define the N-connected
regions in the Cartesian workspace of fully-parallel manipulators,
i.e., the maximal regions where it is possible to execute any
point-to-point motion. The N-connected regions are characterized by
projection, onto the Cartesian space, of the connected components
of the manipulator configuration space defined in the Cartesian
product of the Cartesian space by the joint space. Generalized
Octree models are used for the construction of all spaces. This
study is illustrated with a simple planar fully-parallel
manipulator.
\section{Preliminaries}
Some useful definitions are recalled in this section.
\subsection{Fully-parallel manipulators}
\begin{Def}
A fully-parallel manipulator is a mechanism that includes as many
elementary kinematic chains as the moving platform does admit
degrees of freedom. In addition, every elementary kinematic chain
possesses only one actuated joint (prismatic, pivot or kneecap).
Besides, no segment of an elementary kinematic chain can be linked
to more than two bodies \cite{Merlet:97}.
\label{Definition:Fully_Parallel_Manipulator}
\end{Def}
In this study, kinematic chains, also called ``leg''
\cite{Angeles:97}, will be always independent.
\subsection{Kinematics}
The input vector \negr q (the vector of actuated joint values) is
related to the output vector \negr X (the vector of configuration
of the moving platform) through the following general equation :
\be
        F(\negr X, \negr q)=0
        \protect\label{equation:the_kinematic}
\ee
Vector (\negr X, \negr q) will be called {\em manipulator
configuration} and \negr X is the platform configuration and will
be more simply termed {\em configuration}. Differentiating equation
(\ref{equation:the_kinematic}) with respect to time leads to the
velocity model
\be
     \negr A \negr t + \negr B \dot{\negr q} = 0
\ee
With $\negr t=\left[w, \dot{\negr c} \right]^T$, for planar
manipulators ($w$ is the scalar angular-velocity and $\dot{\negr
c}$ is the two-dimensional velocity vector of the operational point
of the moving platform), $\negr t=\left[\negr w \right]^T$, for
spherical manipulators and $\negr t=\left[\negr w,\dot{\negr
c}\right]^T$, for spatial manipulators ($\dot{\negr c}$ is the
three-dimensional velocity vector and $\dot{\negr w}$ is the
three-dimensional angular velocity-vector of the operational point
of moving platform).
\par
Moreover, \negr A and \negr B are respectively the
direct-kinematics and the inverse-kinematics matrices of the
manipulator. A singularity occurs whenever \negr A or
\negr B, (or both) can no longer be inverted. Three
types of singularities exist \cite{Gosselin:90}:
\beqa
    det(\negr A) &=& 0                       \nonumber \\
    det(\negr B) &=& 0                       \nonumber \\
    det(\negr A) &=& 0 \quad and \quad det(\negr B) = 0  \nonumber
\eeqa
\subsection{Parallel singularities}
Parallel singularities occur when the determinant of the direct
kinematics matrix \negr A vanishes. The corresponding singular
configurations are located inside the Cartesian workspace. They are
particularly undesirable because the manipulator can not resist any
effort and control is lost.
\subsection{Serial singularities}
Serial singularities occur when the determinant of the inverse
kinematics matrix \negr B vanishes. By definition, the
inverse-kinematic matrix is always diagonal: for a manipulator with
$n$ degrees of freedom, the inverse kinematic matrix \negr B can be
written like in equation~(\ref{equation:the_B_matrix}). Each term
$\negr B_{jj}$ is associated with one leg. A serial singularity
occurs whenever at least one of these terms vanishes.
\be
  \negr B = Diag\left[\negr B_{11}, ..., \negr B_{jj}, ..., \negr B_{nn}\right]
  \label{equation:the_B_matrix}
\ee
When the manipulator is in serial singularity, there is a direction
along which no Cartesian velocity can be produced.
\subsection{Postures}
The {\em postures} are defined for fully-parallel manipulators with
multiple inverse kinematic solutions \cite{Chablat:97}. Let $W$ be
the reachable Cartesian workspace, that is, the set of all
reachable configurations of the moving platform (\cite{Kumar:92}
and \cite{Pennock:93}). Let $Q$ be the reachable joint space, that
is, the set of all joint vectors reachable by the actuated joints.
\begin{Def}
For a given configuration \negr X in $W$, a {\em posture} is
defined as a solution to the inverse kinematics of the manipulator.
\end{Def}
 According
to the joint limit values, all postures do not necessarily exist.
Changing posture is equivalent to changing the posture of one or
several legs.
\subsection{Point-to-point trajectories}
There are two major types of tasks to consider~: point-to-point
motions and continuous path tracking. Only point-to-point motions
will be considered in this study.
\begin{Def}
A point-to-point trajectory $T$ is defined by a set of $p$
configurations in the Cartesian workspace~: $T = \{\negr X_1,...,
\negr X_i,
....,  \negr X_p\}$.
\end{Def}
By definition, no path is prescribed between any two configurations
$\negr X_i$ and $\negr X_j$.
\begin{Hyp}
In a point-to-point trajectory, the moving platform can not move
through a parallel singularity.
\end{Hyp}
\par
Although it was shown recently that in some particular cases a
parallel singularity could be crossed \cite{Nenchev:97}, hypothesis
1 is set for the most general cases.
\par
\begin{figure}[!hbt]
    \begin{center}
        \includegraphics[width= 80mm,height= 31.7mm]{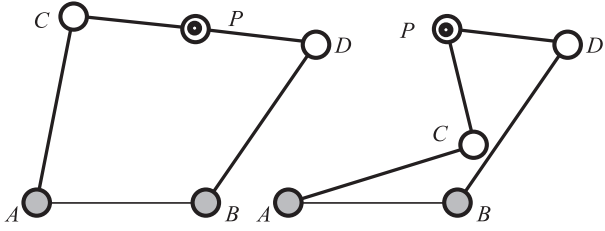}
        \caption{Singular (left) and a regular (right) configurations
        (the actuated joints are $A$ and $B$)}
        \protect\label{figure:singularity_regular_configurations}
    \end{center}
\end{figure}

A point-to-point trajectory $T$ will be feasible if there exists a
continuous path in the Cartesian product of the Cartesian space by
the joint space which does not meet a parallel singularity and
which makes the moving platform pass through all prescribed
configurations $\negr X_i$ of the trajectory $T$.
\begin{Rem}
A fully-parallel manipulator with several inverse kinematic
solutions can change its posture between two prescribed
configurations. Such a manoeuver may enable the manipulator to
avoid a parallel singularity (Figure
\ref{figure:singularity_regular_configurations}). More generally,
the choice of the posture for each configuration $\negr X_i$ of the
trajectory $T$ can be established by any other criteria like
stiffness or cycle time \cite{Chablat:98b}. Note that a change of
posture makes the manipulator run into a serial singularity, which
is not redhibitory for the feasibility of point-to-point
trajectories.
\end{Rem}
\subsection{The generalized octree model}
The quadtree and octree models are hierachical data structures
based on a recursive subdivision of the plane and the space,
respectively \cite{Meagher:81}. There are useful for representing
complex 2-D and 3-D shapes. In this paper, we use a generalization
of this model to dimension $k$, with $k > 3$, the $2^k$-tree
\cite{Chablat:98}. This model is suitable for Boolean operations
like union, difference and intersection. Since this structure has
an implicit adjacency graph, path-connectivity analyses and
trajectory planning can be naturally achieved.
\par
When $k > 3$, it is not possible to represent graphically the
$2^k$-tree. It is necessary to project this structure onto a lower
dimensional space (quadtree or octree). For a n-dof fully-parallel
manipulator, the Cartesian product of the Cartesian space by the
joint space defines generalized octree with dimension $2n$. When
$n=3$ (respectively $n=2$), the projection onto the Cartesian space
and the joint space yields octree models (respectively quadtree
models).
\section{The moveability in the Cartesian workspace}
\begin{Def}
The {\em N-connected regions} of the Cartesian workspace are the
maximal regions where any point-to-point trajectory is feasible.
\end{Def}
For manipulators with multiple inverse and direct kinematic
solutions, it is not possible to study the joint space and the
Cartesian space separately. First, we need to define the {\em
regions of manipulator reachable configurations} in the Cartesian
product of the Cartesian space by the joint space $W.Q$.
\begin{Def}
The regions of manipulator reachable configurations $R_j$ are
defined as the maximal sets in $W . Q$ such that
\end{Def}
\begin{itemize}
\item $R_j \in W . Q$,
\item $R_j$ is connected,
\item $R_j= \{\negr X, \negr q\}$ such that $det(\negr A) \neq 0$
\end{itemize}
In other words, the regions $R_j$ are the sets of all
configurations (\negr X, \negr q) that the manipulator can reach
without meeting a parallel singularity and which can be linked by a
continuous path in $W . Q$.
\begin{Prop}
A trajectory $T=\{\negr X_1,...,\negr X_p\}$ defined in the
Cartesian workspace $W$ is feasible if and only if :
\end{Prop}
\beqa
\left\{\begin{array}{c}
            \forall \negr X \in \{\negr X_1,..., \negr X_p\}\\
            \exists \negr q_i \in Q, \exists R_j
       \end{array}
\right.  ~such~that~(\negr X_i, \negr q_i) \in R_j \nonumber
\eeqa
In other words, for each configuration $\negr X_i$ in $T$, there
exists at least one posture $\negr q_i$ and one region of
manipulator reachable configurations $R_j$ such that the
manipulator configuration $(\negr X_i, \negr q)$ is in $R_j$.
\begin{Pro}
Indeed, if for all configurations  $\negr X_i$, there is one joint
configuration $\negr q_i$ such that $(\negr X_i, \negr q_i) \in
R_j$ then the trajectory is feasible because, by definition, a
region of manipulator reachable configurations is connected and
free of parallel singularity. Conversely, if for a given
configuration $\negr X_i$, it is not possible to find a posture
$\negr q_i$ such that $(\negr X_i, \negr q_i) \in R_j$, then no
continuous, parallel singularity-free path exists in $W . Q$ which
can link the other prescribed configurations.
\end{Pro}
\begin{Theo}
The N-connected regions $W_{N j}$ are the projection ${\Pi}_W$ of
the region of manipulator reachable configurations $R_j$ onto the
Cartesian space :
\end{Theo}
\beqa
W_{N j}= {\Pi}_W R_j \nonumber
\eeqa
\begin{Pro}
This results is a straightforward consequence of the above
proposition.
\end{Pro}
\par
The N-connected regions cannot be used directly for planning
trajectories in the Cartesian workspace since it is necessary to
choose one joint configuration $\negr q$ for each configuration
$\negr X$ of the moving platform such that $(\negr X, \negr q)$ is
included in the same region of manipulator reachable configurations
$R_j$. However, the N-connected regions provide interesting global
information with regard to the performances of a fully-parallel
manipulators because they define the maximal regions of the
Cartesian workspace where it is possible to execute any
point-to-point trajectory.
\par
A consequence of the above theorem is that the Cartesian workspace
$W$ is N-connected if and only if there exists a N-connected region
$W_{N j}$ which is coincident with the Cartesian workspace :
\beqa
W_{N j} = W \nonumber
\eeqa
\section{Example: A Two-DOF fully-parallel manipulator}
\begin{figure}[hbt]
    \begin{center}
        \includegraphics[width= 64.4mm,height= 45 mm]{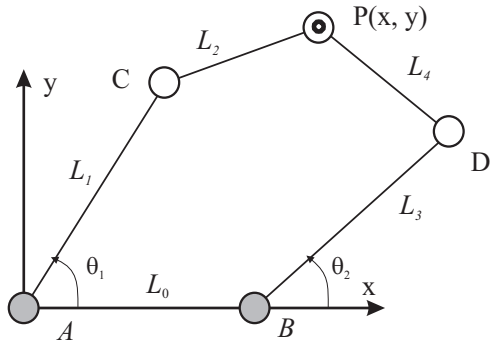}
        \caption{A two-dof fully-parallel manipulator}
        \protect\label{figure:manipulateur_general}
    \end{center}
\end{figure}

For more legibility, a planar manipulator is used as illustrative
example in this paper. This is a five-bar, revolute
($R$)-closed-loop linkage, as displayed in
figure~\ref{figure:manipulateur_general}. The actuated joint
variables are $\theta_1$ and $\theta_2$, while the Output values
are the ($x$, $y$) coordinates of the revolute center $P$. The
passive joints will always be assumed unlimited in this study.
Lengths $L_0$, $L_1$, $L_2$, $L_3$, and $L_4$ define the geometry
of this manipulator entirely. The dimensions are defined in
table \ref{table:dimension} in certain units of length that we need
not specify.
\begin{table}[hbt]
 \begin{center}
 \begin{tabular}{|c|c|c|c|c|c|c|c|c|} \hline
 $\!L_0$&$\!L_1$&$\!L_2$&$\!L_3$&$\!L_4$&
 $\!\theta_{1min}\!\!$&$\!\theta_{1max}\!\!$&
 $\!\theta_{2min}\!\!$&$\!\theta_{2max}\!\!$
 \\ \hline
 7 & 8 & 5 & 8 & 5 & 0 &  $\pi$ &  0 & $\pi$ \\ \hline
 \end{tabular}
 \caption{The dimensions of the RR-RRR studied}
 \label{table:dimension}
\label{table:aspect}
 \end{center}
\end{table}
\par
\begin{figure}[!hbt]
    \begin{center}
        \includegraphics[width= 55mm,height= 39.7mm]{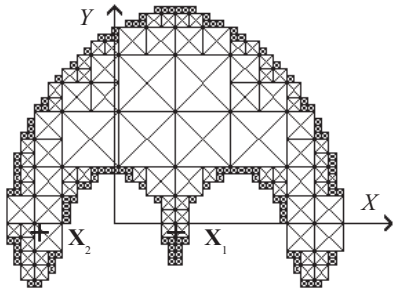}
        \caption{The Cartesian workspace}
        \protect\label{figure:workspace}
    \end{center}
\end{figure}

As shown in table \ref{table:dimension}, the actuated joints are
limited. The Cartesian workspace is shown in figure
\ref{figure:workspace}. We want to know whether this manipulator
can execute any point-to-point motion in the Cartesian workspace.
To answer this question, we need to determine the the N-connected
regions.
\subsection{Singularities}
\begin{figure}[hbt]
    \begin{center}
    \begin{tabular}{ccc}
       \begin{minipage}[t]{35 mm}
         \includegraphics[width= 30mm,height= 23.0mm]{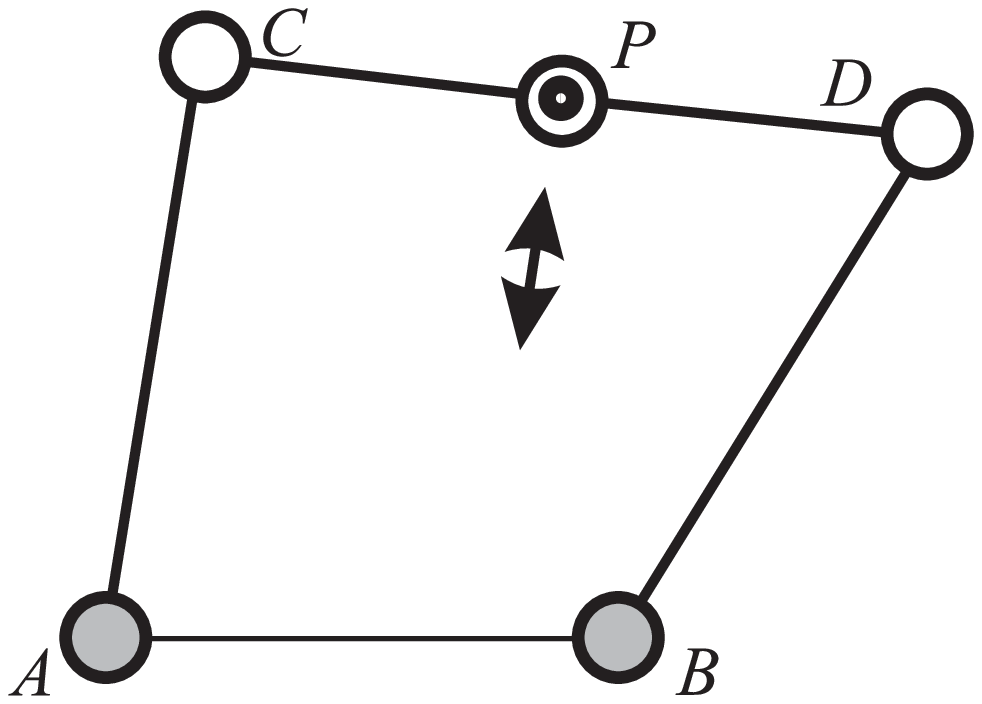}
         \caption{Example of parallel singularity}
         \protect\label{figure:parallel_singularity}
       \end{minipage} &
       \begin{minipage}[t]{35 mm}
        \includegraphics[width= 35mm,height= 29.6mm]{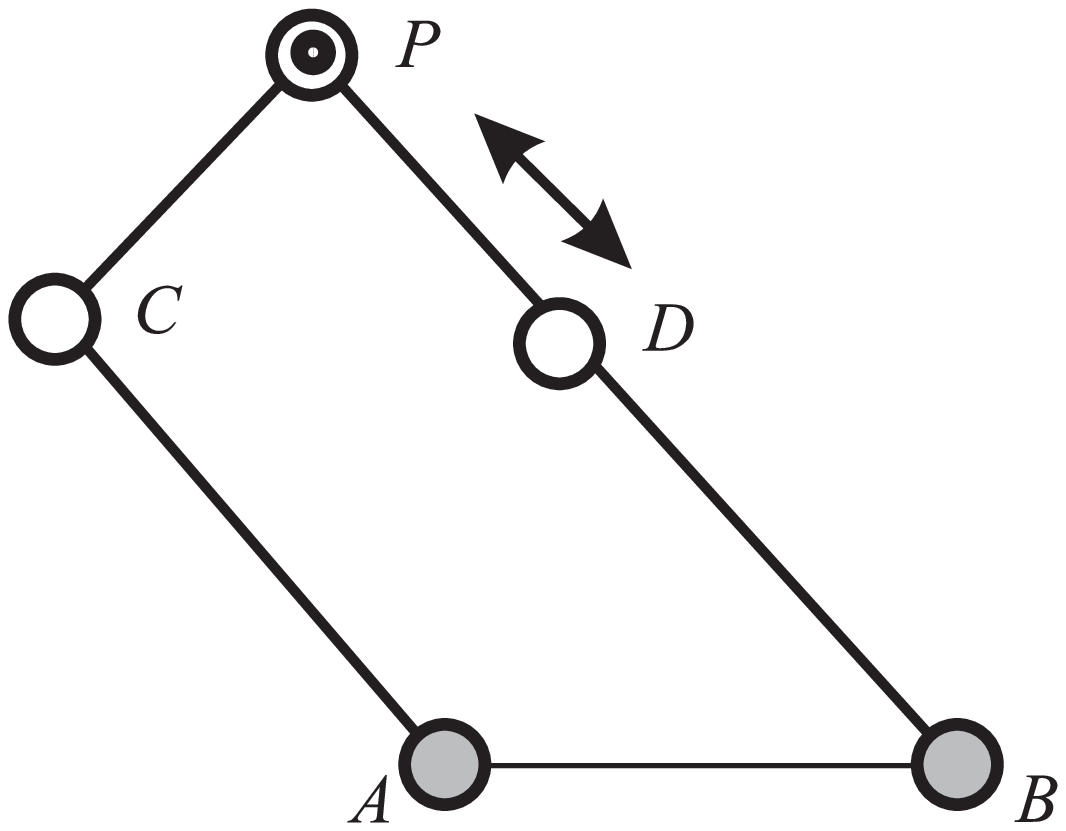}
        \caption{Example of serial singularity}
        \protect\label{figure:serial_singularity}
       \end{minipage}
    \end{tabular}
    \end{center}
\end{figure}

For the manipulator studied, the parallel singularities occur
whenever the points $C$, $D$, and $P$ are aligned (Figure
\ref{figure:parallel_singularity}). Manipulator postures whereby
$\theta_3-\theta_4= k\pi$ denote a singular matrix \negr A, and
hence, define the boundary of the joint space of the manipulator.
For the manipulator at hand, the serial singularities occur
whenever the points $A$, $C$, and $P$ or the points $B$, $D$, and
$P$ are aligned (Figure \ref{figure:serial_singularity}).
Manipulator postures whereby $\theta_3-\theta_1= k\pi$ or
$\theta_4-\theta_2= k\pi$ denote a singular matrix \negr B, and
hence, define the boundary of the Cartesian workspace of the
manipulator.
\subsection{Postures}
\begin{figure}[!hbt]
    \begin{center}
        \includegraphics[width= 65mm,height= 51.6mm]{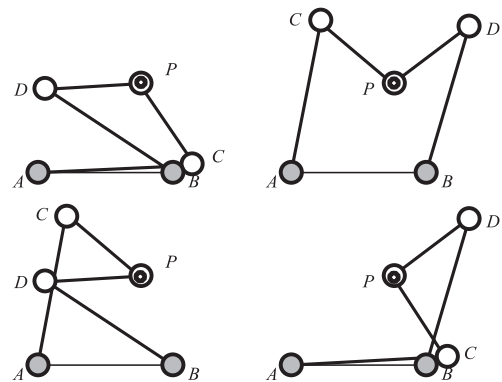}
        \caption{The four postures}
        \protect\label{figure:working_mode}
    \end{center}
\end{figure}

The manipulator under study has four postures, as depicted in
figure~\ref{figure:working_mode}. According to the posture, the
parallel singularity locus changes in the Cartesian workspace, as
already shown in
figure~\ref{figure:singularity_regular_configurations}.
\subsection{The N-connected regions}
\begin{figure}[hbt]
  \begin{center}
    \includegraphics[width= 72mm,height= 55mm]{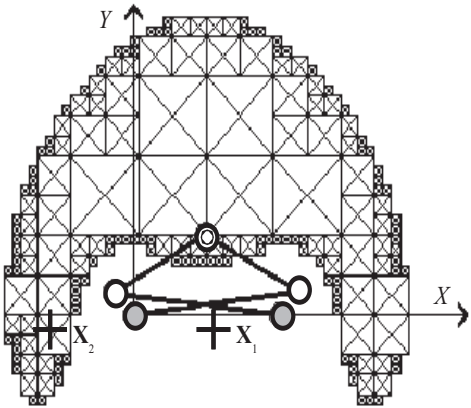}
    \caption{The first N-connected region of the Cartesian workspace when $0.0 \le \theta_1, \theta_2 \le \pi$}
    \protect\label{figure:First_N_connected_region}
  \end{center}
\end{figure}

\begin{figure}[hbt]
  \begin{center}
    \includegraphics[width= 72mm,height= 55mm]{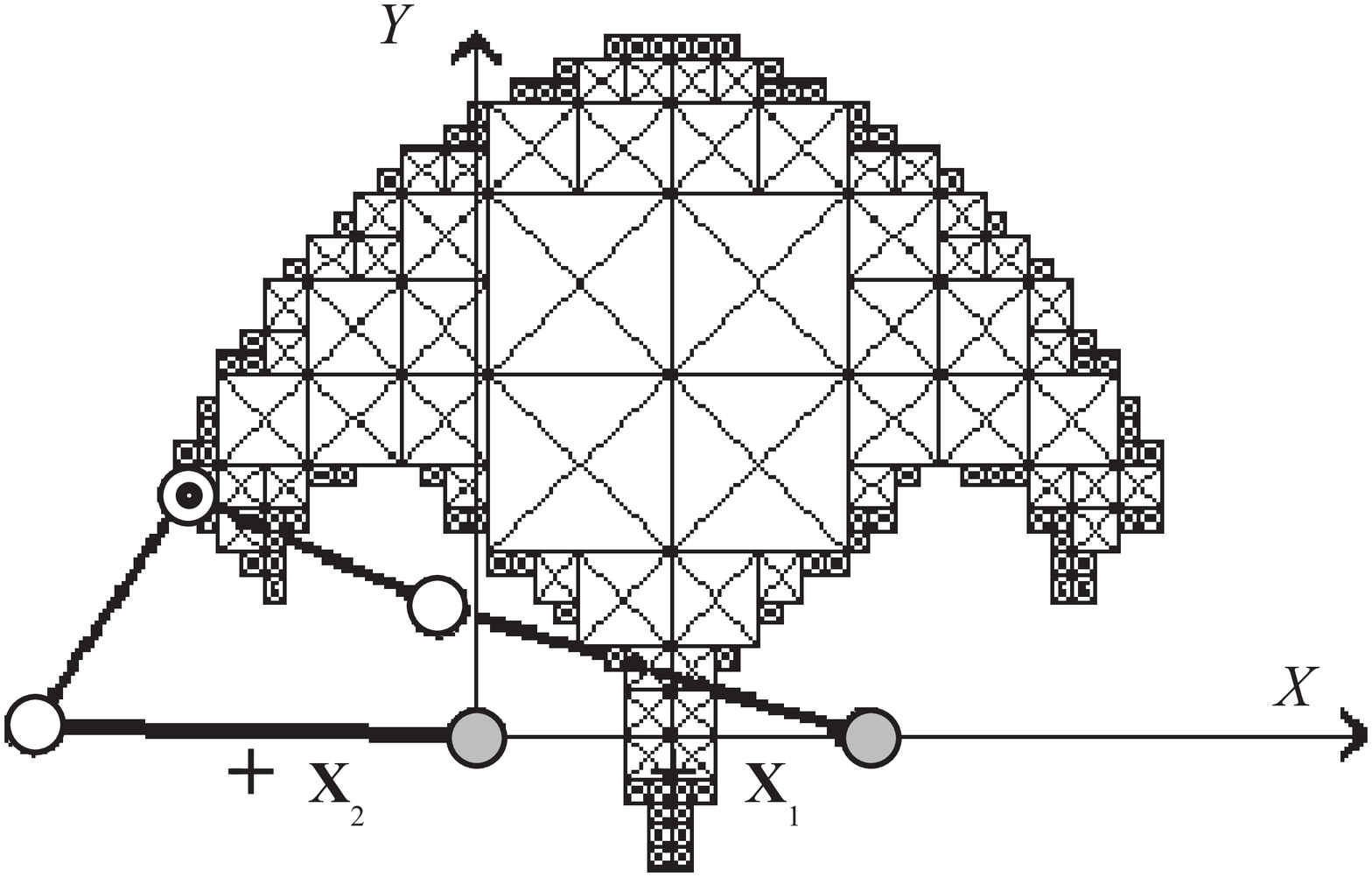}
    \caption{The second N-connected region of the Cartesian workspace when $0.0 \le \theta_1, \theta_2 \le \pi$}
    \protect\label{figure:Second_N_connected_region}
  \end{center}
\end{figure}

It turns out that the Cartesian workspace of the manipulator at
hand is not N-connected, e.g. the manipulator cannot move its
platform between any set of configurations in the Cartesian
workspace. In effect, due to the existence of limits on the
actuated joints, not all postures are accessible for any
configuration in the Cartesian workspace. Thus, the manipulator may
loose its ability to avoid a parallel singularity when moving from
one configuration to another. This is what happens between points
$\negr X_1$ and $\negr X_2$ (Figure \ref{figure:workspace}). These
two points cannot be linked by the manipulator although they lie in
the Cartesian workspace which is connected in the mathematical
sense (path-connected) but not N-connected. In fact, there are two
separate N-connected regions which do not coincide with the
Cartesian workspace and the two points do not belong to the same
N-connected region (Figures \ref{figure:First_N_connected_region}
and \ref{figure:Second_N_connected_region}).
\par
Physically, any attempt in moving the point P from $\negr X_1$ to
$\negr X_2$ will cause the manipulator either cross a parallel
singularity or reach a joint limit.
\par
\begin{figure}[hbt]
    \begin{center}
    \begin{tabular}{cc}
       \begin{minipage}[t]{42 mm}
           \centerline{\hbox{\includegraphics[width= 40.7mm,height= 29.4mm]{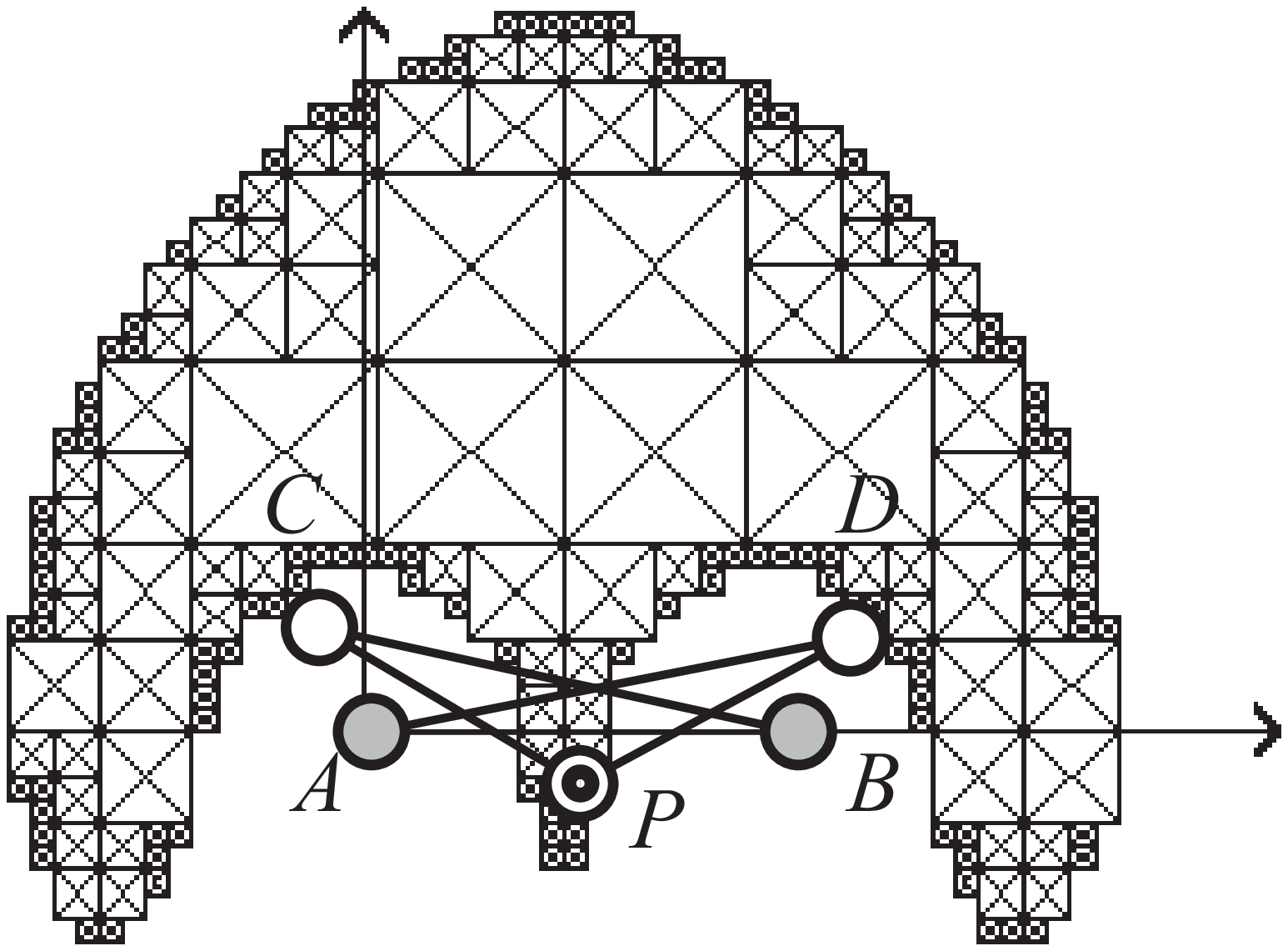}}}
           \begin{center}
             (a)
           \end{center}
       \end{minipage} &
       \begin{minipage}[t]{42 mm}
           \centerline{\hbox{\includegraphics[width= 40.7mm,height= 29.4mm]{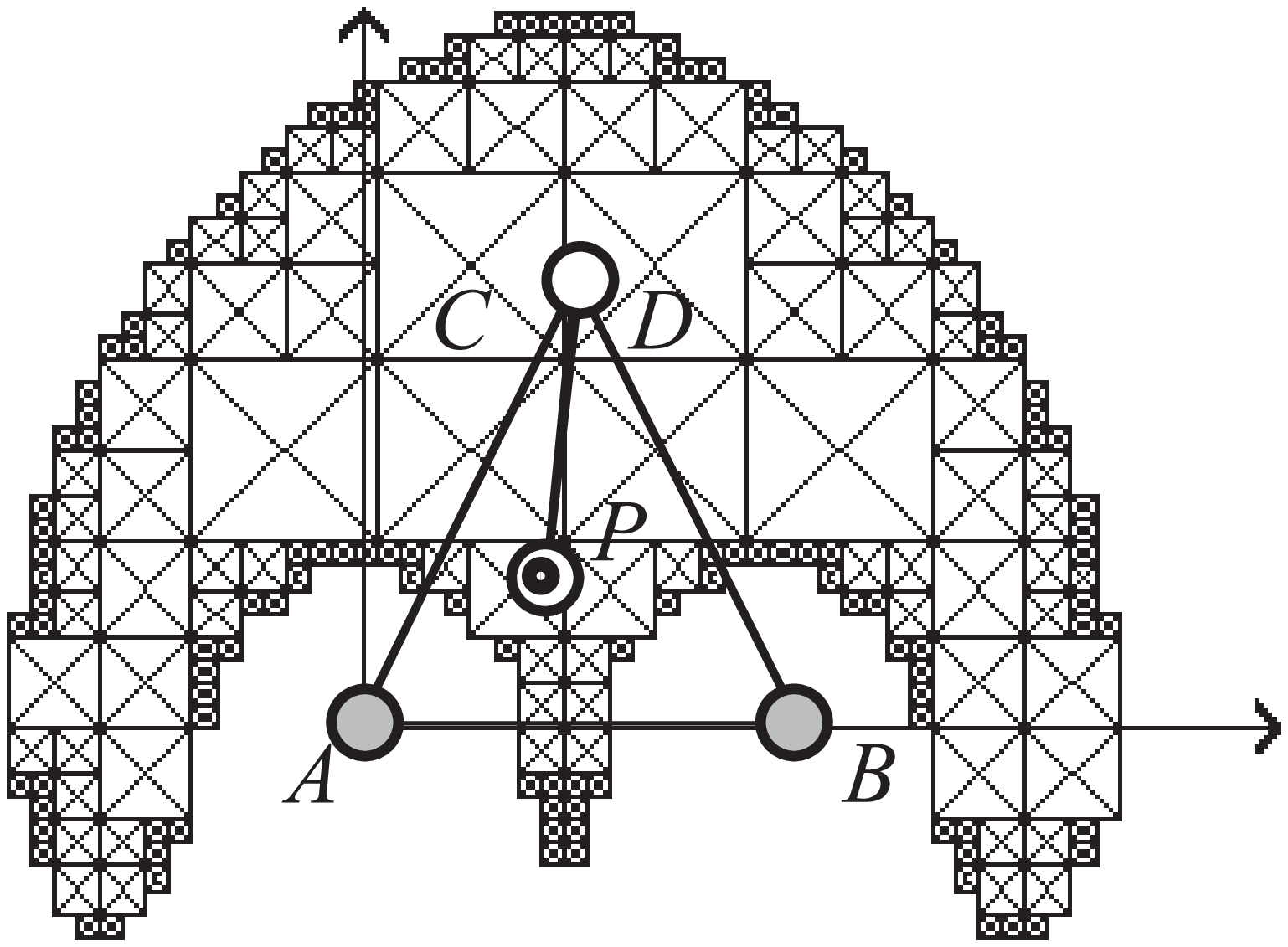}}}
           \begin{center}
             (b)
           \end{center}
       \end{minipage} \\
       \begin{minipage}[t]{42 mm}
           \centerline{\hbox{\includegraphics[width= 40.7mm,height= 29.4mm]{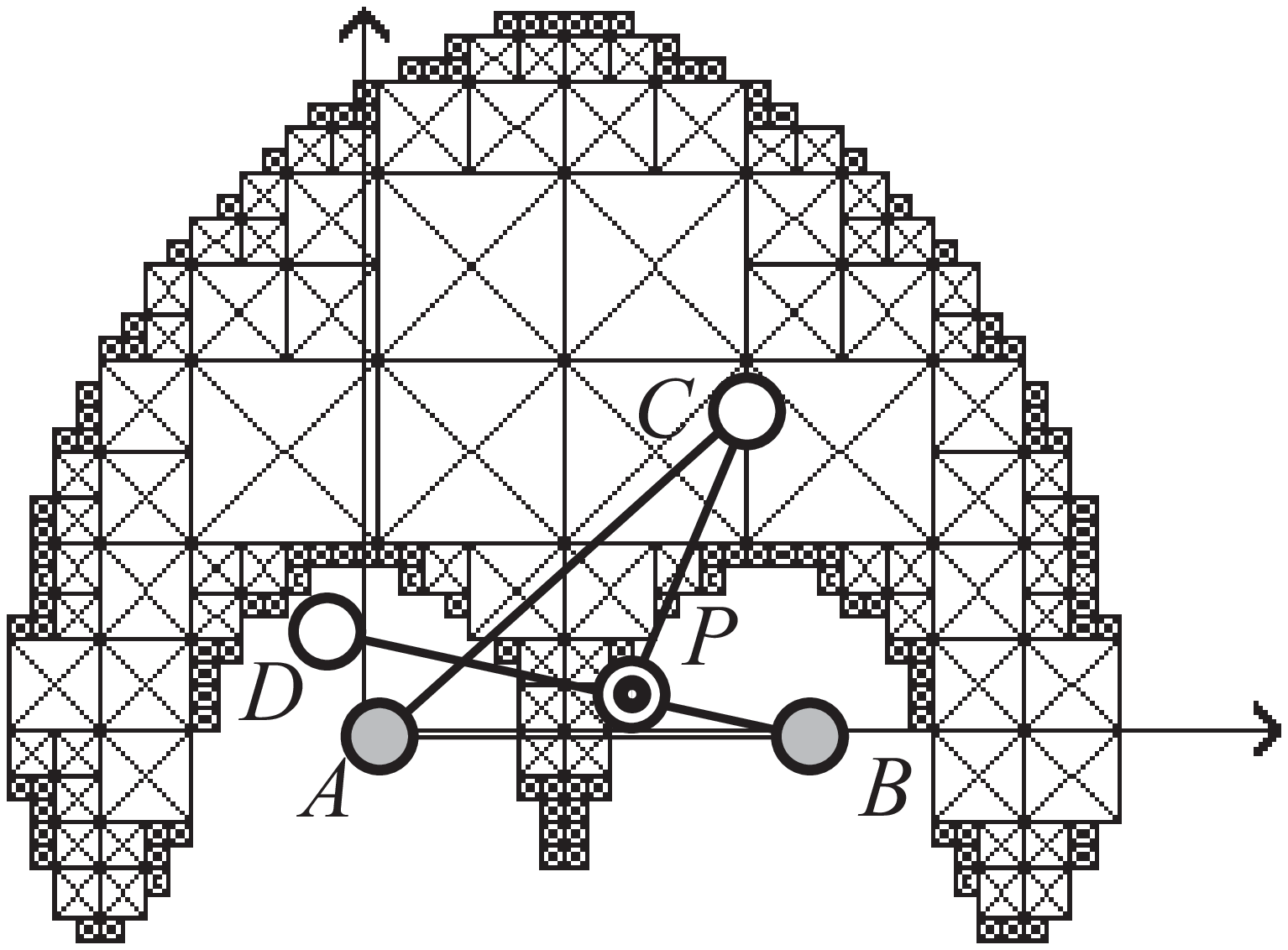}}}
           \begin{center}
             (c)
           \end{center}
       \end{minipage} &
       \begin{minipage}[t]{42 mm}
           \centerline{\hbox{\includegraphics[width= 40.7mm,height= 29.4mm]{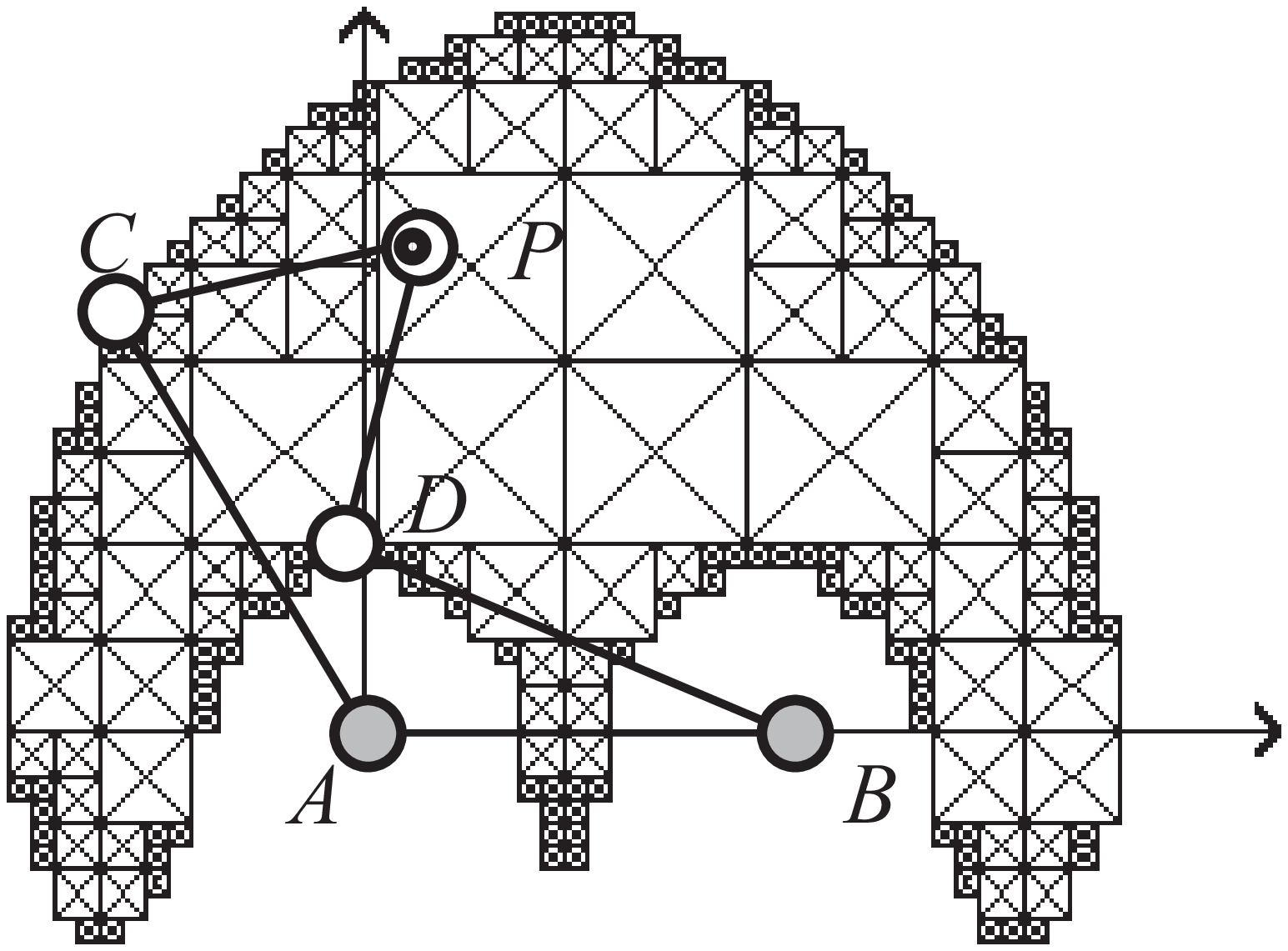}}}
           \begin{center}
             (d)
           \end{center}
       \end{minipage} \\
       \begin{minipage}[t]{42 mm}
           \centerline{\hbox{\includegraphics[width= 40.7mm,height= 29.4mm]{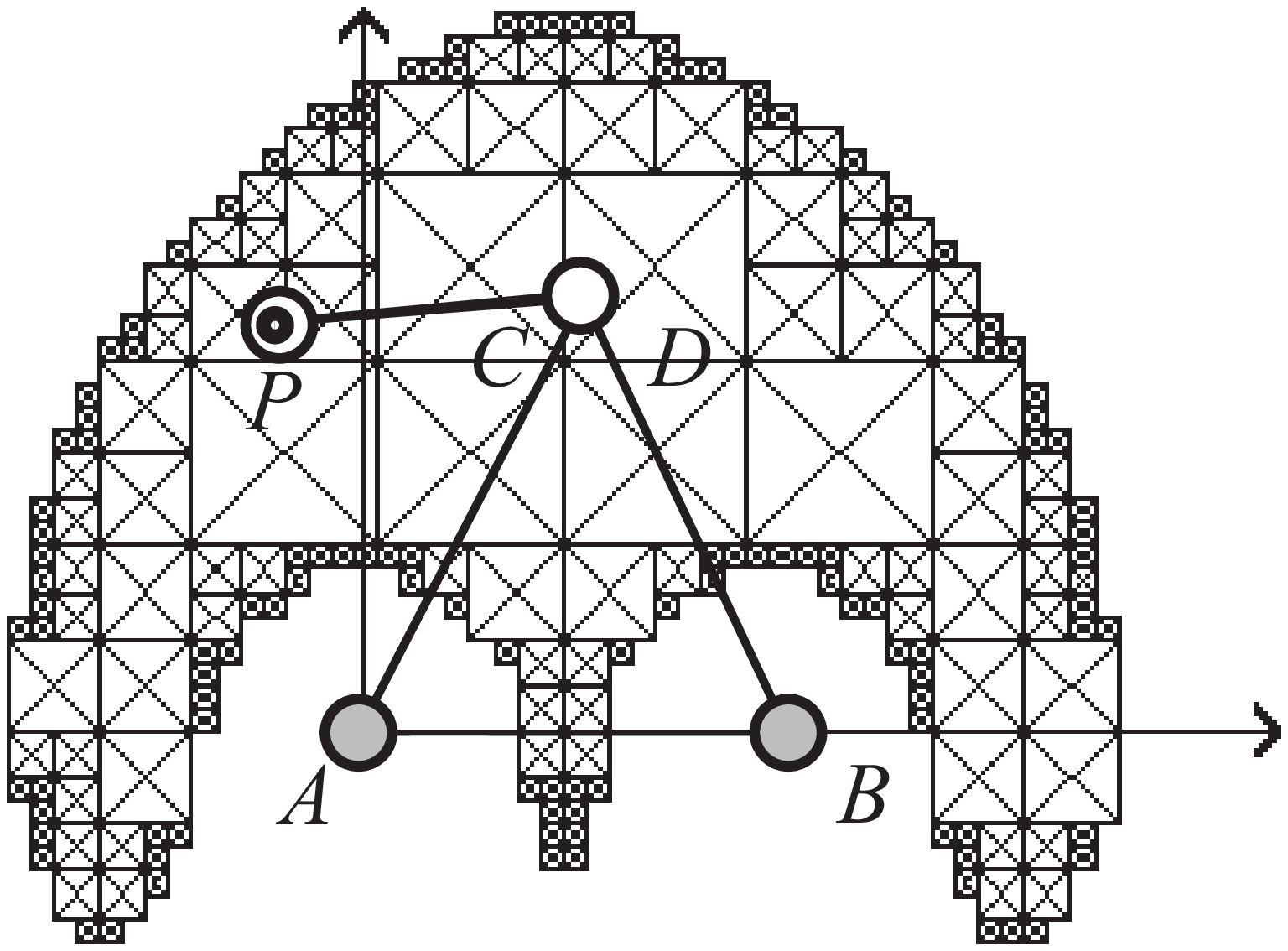}}}
           \begin{center}
             (e)
           \end{center}
       \end{minipage} &
       \begin{minipage}[t]{42 mm}
           \centerline{\hbox{\includegraphics[width= 40.7mm,height= 29.4mm]{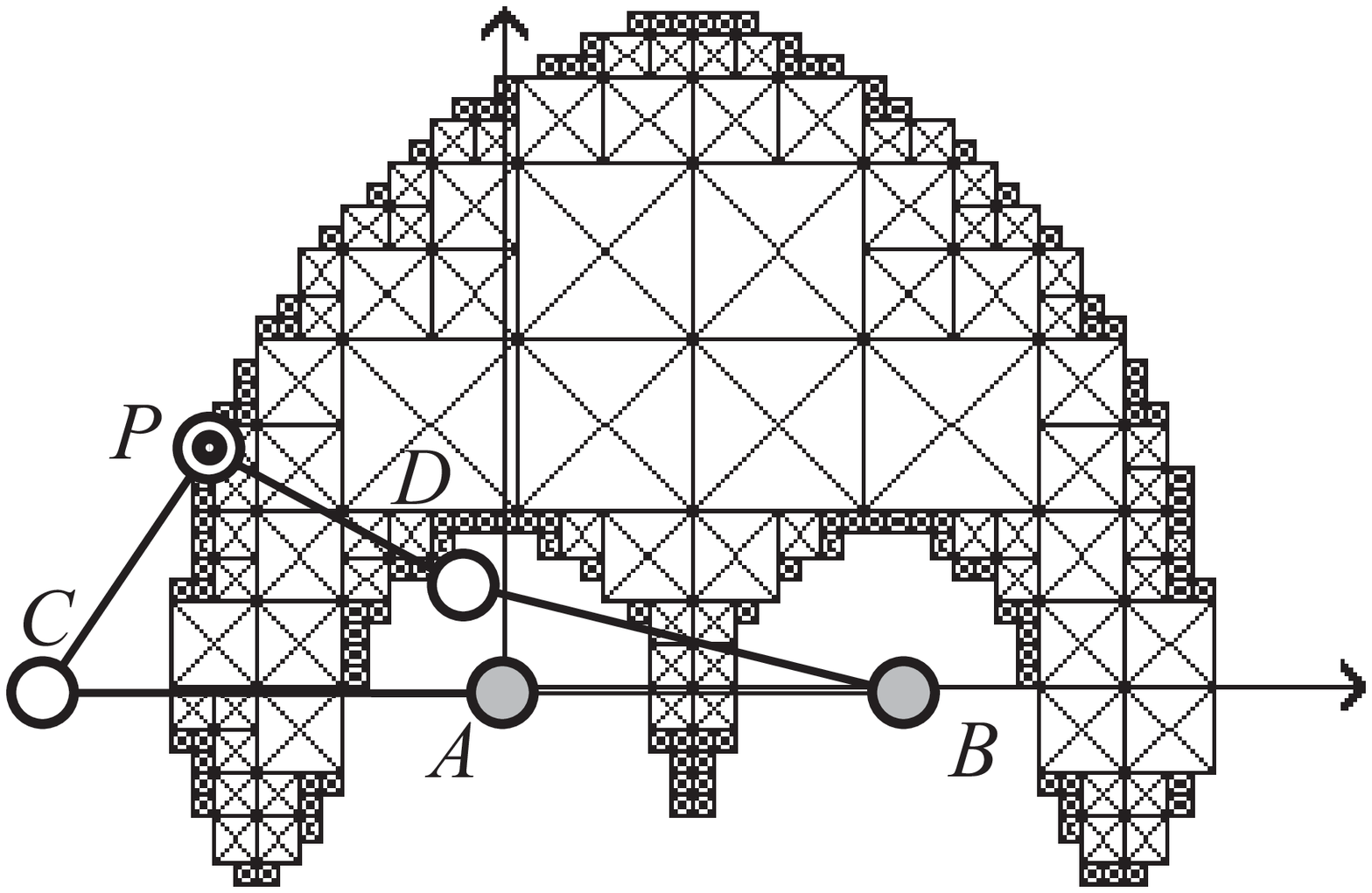}}}
           \begin{center}
             (f)
           \end{center}
       \end{minipage}
    \end{tabular}
    \caption{Moving from $\negr X_1$ to $\negr X_4$}
    \protect\label{figure:Path}
    \end{center}
\end{figure}

In effect, point $\negr X_1$ is accessible only in the manipulator
configuration shown in figure \ref{figure:Path}a because of the
joint limits. When moving towards point $\negr X_4$, the
manipulator cannot remain in its initial posture because it would
meet a parallel singularity (Figure \ref{figure:Path}b). Thus, it
must change its posture, let say at $\negr X_3$ (Figure
\ref{figure:Path}c). The only new posture which can be chosen is
the one depicted in figure \ref{figure:Path}d because any other
posture would make the manipulator meet a parallel singularity
(Figure \ref{figure:Path}e). Then, it is apparent that the
manipulator cannot reach $\negr X_1$ from $\negr X_4$ since joint A
attains its limits (figure \ref{figure:Path}f).
\par
\begin{figure}[t]
    \begin{center}
        \includegraphics[width= 70mm,height= 70mm]{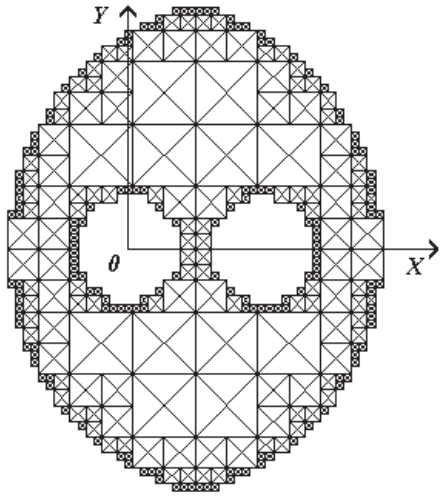}
        \protect\center\caption{The N-connected regions of the Cartesian workspace when $-\pi \le \theta_1, \theta_2 \le \pi$}
        \protect\label{figure:N_connected_region}
    \end{center}
\end{figure}

If we change the values of the joint limits ($\theta_{1 min} =
\theta_{2 min} = -\pi$), the Cartesian workspace is now N-connected since
the computed N-connected regions are coincident with the Cartesian
workspace (Figure \ref{figure:N_connected_region}). In effect, it
can be verified in this case that for every configuration of the
moving platform, there are four postures which define two regions
of accessible configurations whose projection onto the Cartesian
space yields the full Cartesian workspace.
\section{Conclusions}
The aim of this paper was the characterization of the N-connected
regions in the Cartesian workspace of fully-parallel manipulators,
i.e. the regions of feasible point-to-point trajectories. The word
feasible means that the manipulator should be able to move between
all prescribed configurations while never meeting a parallel
singularity. The manipulators considered in this study have
multiple solutions to their direct and inverse kinematics. The
N-connected regions were defined by first determining the maximum
path-connected, parallel singularity-free regions in the Cartesian
product of the Cartesian workspace by the joint space. The
projection of these regions onto the Cartesian workspace were shown
to define the N-connected regions.
\par
The N-connectivity analysis of the Cartesian workspace is of high
interest for the evaluation of manipulator global performances as
well as for off-line task programming.
\par
Further research work is being conducted by the authors to take
into account the collisions and to characterize the maximum regions
of the Cartesian workspace where the manipulator can track any
continuous trajectory.


\begin{thebibliography}{}
\bibitem[Wenger, 91]{Wenger:91}
Wenger, Ph., Chedmail, P.
\newblock ``Ability of a Robot to Travel Through its Free Workspace''
\newblock The International Journal of Robotic Research, Vol. 10:3, June 1991.
\bibitem[Chablat, 98a]{Chablat:98}
Chablat, D.
\newblock ``Domaines d'unicit\'e et parcourabilit\'e pour
les manipulateurs pleinement parall\`eles''
\newblock PhD thesis, Nantes, November 1998.
\bibitem[Chablat, 97]{Chablat:97}
Chablat, D. and Wenger, Ph.
\newblock``Working modes and aspects in fully-parallel manipulators''
\newblock Proceeding IEEE International Conference of Robotic and Automation,
pp. 1964-1969, May 1998.
\bibitem[Merlet, 97]{Merlet:97}
Merlet, J-P.
\newblock ``Les robots parall\`eles''
\newblock HERMES, seconde \'edition, Paris, 1997.
\bibitem[Angeles, 97]{Angeles:97}
Angeles, J.
\newblock ``Fundamentals of Robotic Mechanical Systems''
\newblock SPRINGER 97.
\bibitem[Gosselin, 90]{Gosselin:90}
Gosselin, C.\ and Angeles, J.
\newblock ``Singularity analysis of closed-loop kinematic chains''
\newblock IEEE Transactions On Robotics And Automation, Vol.~6, No.~3,
June 1990.
\bibitem[Kumar, 92]{Kumar:92}
Kumar V.
\newblock ``Characterization of workspaces of parallel manipulators''
\newblock ASME J. Mechanical Design, Vol. 114, pp 368-375, 1992.
\bibitem[Pennock, 93]{Pennock:93}
Pennock, G.R. and Kassner, D.J.
\newblock ``The workspace of a general geometry planar
three-degree-of-freedom platform-type manipulator''
\newblock ASME J. Mechanical Design, Vol. 115, pp 269-276, 1993.
\bibitem[Nenchev, 97]{Nenchev:97}
Nenchev, D.N., Bhattacharya, S., and Uchiyama, M.,
\newblock ``Dynamic Analysis of Parallel Manipulators under the
Singularity-Consistent Parameterization''
\newblock Robotica, Vol.~15, pp. 375-384. 1997.
\bibitem[Chablat, 98b]{Chablat:98b}
Chablat, D., Wenger, Ph. , Angeles, J.
\newblock {\em ``The isoconditioning Loci of A Class of Closed-Chain Manipulators''}
\newblock Proceeding IEEE International Conference of Robotic and Automation, pp. 1970-1975,
May 1998.
\bibitem[Meagher, 81]{Meagher:81}
Meagher, D.
\newblock ``Geometric Modelling using Octree Encoding''
\newblock Technical Report IPL-TR-81-005, Image Processing Laboratory,
Rensselaer Polytechnic Institute, Troy, New York 12181, 1981.
\end{thebibliography}
\end{document}